# Jeffrey's rule of conditioning generalized to belief functions.

Philippe SMETS[1]
IRIDIA, Université Libre de Bruxelles
50 av. F. Roosevelt, CP194/6. B-1050 Brussels, Belgium.

**Abstract:**
Jeffrey's rule of conditioning has been proposed in order to revise a probability measure by another probability function. We generalize it within the framework of the models based on belief functions. We show that several forms of Jeffrey's conditionings can be defined that correspond to the geometrical rule of conditioning and to Dempster's rule of conditioning, respectively.

## 1. Jeffrey's rule in probability theory.

In probability theory conditioning on an event is classically obtained by the application of Bayes' rule. Let $(\Omega, \mathcal{A}, P)$ be a probability space where $P(A)$ is the probability of the event $A \in \mathcal{A}$ where $\mathcal{A}$ is a Boolean algebra defined on a finite[2] set $\Omega$. $P(A)$ quantified the degree of belief or the objective probability, depending on the interpretation given to the probability measure, that a particular arbitrary element $\varpi$ of $\Omega$ which is not a priori located in any of the sets of $\mathcal{A}$ belongs to a particular set $A \in \mathcal{A}$. Suppose it is known that $\varpi$ belongs to $B \in \mathcal{A}$ and $P(B) > 0$. The probability measure P must be updated into $P_B$ that quantifies the same event as previously but after taking in due consideration the knowledge that $\varpi \in B$. $P_B$ is obtained by Bayes' rule of conditioning:

$$P_B(A) = P(A \mid B) = \frac{P(A \cap B)}{P(B)}.$$

This rule can be obtained by requiring that:

**B1:** $\forall B \in \mathcal{A}, \; P_B(B) = 1$
**B2:** $\forall B \in \mathcal{A}, \forall X, Y \in \mathcal{A}$ such that $X, Y \subseteq B$,
$$\frac{P_B(X)}{P_B(Y)} = \frac{P(X)}{P(Y)} \quad \text{if } P(Y) > 0$$
and $\quad P_B(Y) = 0 \quad \text{if } P(Y) = 0.$

Jeffrey (1965) has considered a generalization of Bayes' rule where the updating information does not correspond to the knowledge that $\varpi \in B$, which implies that the updated probability should give a probability 1 to B (see B1), but to the weaker requirement that there is a new probability measure on a sub-algebra $\mathcal{B}$ of $\mathcal{A}$, and the updated probability should give these probability weights to the elements of $\mathcal{B}$.

Let $(\Omega, \mathcal{A}, P_1)$ be the initial probability space. Let $\mathcal{B}$ be a subalgebra of $\mathcal{A}$. Let $(\Omega, \mathcal{B}, P_2)$ be the new probability space. $P_2$ corresponds to the updating information that in fact $\varpi$ belongs to $X \in \mathcal{B}$ with probability $P_2(X)$. The problem is to update $P_1$ into some $P_3$ defined on $\mathcal{A}$ and such that $P_3(X) = P_2(X)$ $\forall X \in \mathcal{B}$. $P_3$ is the result of revising $P_1$ by the probability measure $P_2$, adopting $P_2$ wherever $P_2$ is defined.

Bayes' rule corresponds to the limiting case where one of the atoms[3] of $\mathcal{B}$ receives a probability one, i.e. there is one atom B of $\mathcal{B}$ with $P_2(B) = 1$. Bayes' rule tells nothing on how to build $P_3$ from $P_1$ and $P_2$ in the generalized case. Let $\mathbb{B} = \{B_1, B_2 ... B_n\}$ be the set of atoms on $\mathcal{B}$. Jeffrey requires that $P_3$ should satisfy two requirements:

**R1:** $\forall X \in \mathcal{B}, \; P_3(X) = P_2(X)$
**R2:** $\forall B \in \mathbb{B}, \forall X, Y \in \mathcal{A}$ such that $X, Y \subseteq B$,
$$\frac{P_3(X)}{P_3(Y)} = \frac{P_1(X)}{P_1(Y)} \quad \text{if } P_1(Y) > 0$$
and $\quad P_3(Y) = 0 \quad \text{if } P_1(Y) = 0.$

These two requirements lead to Jeffrey's rule of conditioning:

$$\begin{aligned} P_3(A) &= \sum_{B \in \mathbb{B}} P_1(A|B) \, P_2(B) \\ &= \sum_{B \in \mathbb{B}} \frac{P_1(A \cap B)}{P_1(B)} P_2(B) \end{aligned}$$

where $P_1(A|B) = 0$ if $P_1(B) = 0$.

---

[1] This work has been partially funded by the CEC-ESPRIT III Basic Research Project 6156 (DRUMS II), and the Communauté Française de Belgique, ARC 92/97-160 (BELON).

[2] For simplicity sake, we only consider finite spaces.

[3] An atom of an algebra is a non empty element of the algebra which intersection with the other elements of the algebra equals itself or is empty. When the algebra is a power set, the atoms are the singletons.

In this paper, we generalize this rule in a context where beliefs are quantified by belief functions. Before introducing the generalized Jeffrey rules of conditioning, we present the concepts of revision and focusing. Depending on the type of conditioning event, revision leads to several rules, in particular the geometrical rule and Dempster's rule of conditioning. Both admit a generalization in the spirit of Jeffrey's rules of conditioning. Multiplicity of conditioning rules was already presented in Smets (1991). Previous attempts to generalize Jeffrey's rules of conditioning are discussed in the conclusions.

## 2. Revision versus focusing.

1. Dubois and Prade (1991a) have introduced beautifully the difference between two types of conditioning :

**Case 1.** A die has been tossed. You assess the probability that the outcome is 'Six'. Then a reliable witness says that the outcome is an even number. How do you update the probability that the outcome is 'six' taking in due consideration the new piece of information.

**Case 2.** Hundred dice have been tossed. You assess the proportion of 'six'. Then you decide to focus your interest on the dice with an even outcome. How do you compute the proportion of 'six' among the dice with an even outcome.

Case 1 corresponds to a revision[4] as the probability is modified to take in account a new piece of information.
Case 2 corresponds to a focusing : no new piece of information is introduced, we just consider another reference class by focusing our attention on a given subset of the original set.

In probability theory, the distinction is more conceptual than practical as both cases are solved by Bayes' rule of conditioning. It might explain the lack of interest for such a distinction. We restrict ourselves to the revision case and study what happens when probability measures are replaced by belief functions like in the Transferable Belief Model (Smets, 1988, Smets and Kennes, 1990), in Dempster-Shafer Model (Shafer, 1976a) and in the Hints Model (Kohlas and Monney, 1990).

2. We consider first the probability of provability (deductibility) approach (Pearl 1988, Ruspini 1986). It is not different from the original Dempsterian approach (Dempster 1967) but it provides a nice framework. It can be described as follows.

Let $\mathcal{H}$ be a finite Boolean algebra of propositions. These propositions are called the hypothesis. Let $\mathcal{L}$ be another finite Boolean algebra of propositions. We assume that for every hypothesis $H \in \mathcal{H}$ there is a set $\mathbb{L}(H) = \{L_i: L_i \in \mathcal{L}, i=1, 2..r\}$ of propositions $L_i$ provable under H. Let $M(H) = L_1 \& L_2 ... \& L_r$ be the conjunction of all these propositions in $\mathbb{L}(H)$. $M(H) \in \mathcal{L}$ as $\mathcal{L}$ is a Boolean algebra. M is a function from $\mathcal{H}$ to $\mathcal{L}$.

So: $\quad \forall H \in \mathcal{H}, \exists M(H) \in \mathcal{L}$ such that
$H \vdash L$ for every $L \in \mathcal{L}$ such that $M(H) \vdash L$.

Suppose there is a probability measure $P_{\mathcal{H}}: 2^{\mathcal{H}} \to [0,1]$ on $2^{\mathcal{H}}$ and let $p_{\mathcal{H}}: \mathcal{H} \to [0,1]$ be the related probability function on $\mathcal{H}$ with $p_{\mathcal{H}}(H) = P_{\mathcal{H}}(\{H\})$ for every $H \in \mathcal{H}$. Note that $\mathcal{H}$ is already an algebra, usually the power set of some set. As $\bot \in \mathcal{H}$, $p_{\mathcal{H}}(\bot)$ may be positive.

Given the function $M: \mathcal{H} \to \mathcal{L}$, we can define the probability $P_{\mathcal{L}}$ that $L \in \mathcal{L}$ is provable (deductible) and $\neg L$ is not provable, denoted $P_{\mathcal{L}}(\mapsto L)$. We use the symbol $\mapsto$ in $H \mapsto L$ to mean $H \vdash L$ and $H \not\vdash \neg L$, and in $P_{\mathcal{L}}(\mapsto L)$ to enhance the fact that those H that would also prove $\neg L$ are not included in it. $P_{\mathcal{L}}(\mapsto L)$ is the probability that an hypothesis selected randomly in $\mathcal{H}$ (according to the probability measure $P_{\mathcal{H}}$) proves L and does not prove $\neg L$ (thus eliminating the hypothesis equivalent to the contradiction, denoted $\bot$):

$$P_{\mathcal{L}}(\mapsto L) = P_{\mathcal{H}}(\{H : H \in \mathcal{H}, H \mapsto L\}) = \sum_{H \mapsto L} p_{\mathcal{H}}(H)$$

Let bel: $\mathcal{L} \to [0,1]$ be the belief function[5] on $\mathcal{L}$ induced by a basic belief assignment m on $\mathcal{L}$. By definition bel(L) is the sum of the basic belief masses given to the propositions that imply L without implying $\neg L$ (thus excluding $\bot$):

$$\text{bel}(L) = \sum_{X: X \in \mathcal{L}, X \mapsto L} m(X)$$

It can be shown that $P_{\mathcal{L}}(\mapsto L)$, $L \in \mathcal{L}$ is equal to the belief function $\text{bel}_{\mathcal{L}}$ on $\mathcal{L}$ induced by the basic belief assignment $m_{\mathcal{L}}: \mathcal{L} \to [0,1]$ with:

$$m_{\mathcal{L}}(L) = \sum_{H: M(H)=L} p_{\mathcal{H}}(H)$$

and $m_{\mathcal{L}}(L) = 0$ if the sum is taken over an empty set.

---

[4] In Dubois and Prade (1991), they called it an 'updating' but they prefer to call it now (Dubois and Prade 1992) a revision in harmony with the Alchouroron, Gärdenfors and Makinson approach (Gärdenfors, 1988) where revision concerns the beliefs held by an agent. They reserve 'updating' for the case considered by Katsuno and Mendelzon (1991) that concerns the update of an evolving world.

[5] bel is an unnormalized belief function as we do not require $m(\bot)=0$ (Smets 1988, 1992a). When $m(\bot)>0$, then $\text{bel}(\top) = \text{pl}(\top) < 1$.

502 Smets

One has:
$$P_{\mathscr{L}}(\vdash L) = \sum_{H: H \vdash L} p_{\mathscr{H}}(H) = \sum_{H: M(H) \vdash L} p_{\mathscr{H}}(M(H))$$
$$= \sum_{L_i \vdash L} \sum_{H: M(H) = L_i} p_{\mathscr{H}}(H) = \sum_{L_i \vdash L} m_{\mathscr{L}}(L_i)$$

So:  $P_{\mathscr{L}}(\vdash L) = \text{bel}_{\mathscr{L}}(L)$.

Similarly the plausibility function $\text{pl}_{\mathscr{L}}$ is:

$$\text{pl}_{\mathscr{L}}(L) = \text{bel}_{\mathscr{L}}(\top) - \text{bel}_{\mathscr{L}}(\neg L) = \sum_{X \& L \neq \bot} m_{\mathscr{L}}(X)$$

where $\top$ is the maximal element of $\mathscr{L}$.

This model is not different from Dempster's model (Dempster 1967) and Shafer's translator model (Shafer and Tversky 1985, Dubois et al. 1991). Both models consider an X domain (the translator-source domain) endowed with a probability measure, a Y domain (the message-data domain) and a one-to-many mapping from X to Y. The $\mathscr{H}$ space corresponds to the X domain, the $\mathscr{L}$ to the Y, the M mapping to the one-to-many mapping, and $P_{\mathscr{H}}$ to the probability on X space.

3. We proceed by considering two revision processes that correspond to some data-conditioning and some source-conditioning, i.e. conditioning on an information relative to the data or the source (Moral 1993)

3.1. The data-conditioning fits to the scenario where we learn that a particular proposition L* of $\mathscr{L}$ is true. In that case, the hypothesis H that was proving all the propositions in $\mathscr{L}$ proved by M(H) now proves all propositions in $\mathscr{L}$ proved by M(H)&L*. The basic belief assignment $m_{\mathscr{L}}$ is updated into $m_{\mathscr{L}}^*$ with:

$$m_{\mathscr{L}}^*(L) = \sum_{H: M(H) \& L^* = L} p_{\mathscr{H}}(H)$$

As $\forall L \vdash L^*, L \& L^* = L$, one has:

$$\text{pl}_{\mathscr{L}}^*(L) = \sum_{X \& L \neq \bot} m_{\mathscr{L}}^*(X)$$
$$= \sum_{X: X \& L \neq \bot} \sum_{H: M(H) \& L^* = X} p_{\mathscr{H}}(H)$$
$$= \sum_{H: M(H) \& L \& L^* \neq \bot} p_{\mathscr{H}}(H) = \text{pl}_{\mathscr{L}}(L)$$

This relation corresponds to the unnormalized rule of conditioning (Dempster's rule of conditioning without normalization (Smets 1993a)). Normalization is achieved by further conditioning on L being not equivalent to a contradiction.

3.2. The source-conditioning fits with the following scenario. Given $L^* \in \mathscr{L}$, we consider only those hypothesis H that prove L* (without proving ¬L*) and ask what is then the probability that L is provable for all $L \in \mathscr{L}$ that prove L*. Therefore we restrict our attention to those $H \in \mathscr{H}$ such that $H \vdash L^*$, $H \nvdash \neg L^*$. Let $P_{\mathscr{L}}^{**}(\vdash L)$ be the probability that L is provable by one of those hypothesis that prove L*. Then:

$$P_{\mathscr{L}}^{**}(\vdash L) = P_{\mathscr{H}}(H: H \vdash L \mid H \vdash L^*)$$
$$= \frac{P_{\mathscr{H}}(H: H \vdash L \& L^*)}{P_{\mathscr{H}}(H: H \vdash L^*)}$$
$$= \frac{P_{\mathscr{H}}(H: H \vdash L)}{P_{\mathscr{H}}(H: H \vdash L^*)}$$
$$= \frac{\text{bel}_{\mathscr{L}}(L \& L^*)}{\text{bel}_{\mathscr{L}}(L^*)}$$

This is known as the geometrical rule of conditioning (Shafer 1976b, Suppes and Zanotti, 1977).

3.3. When M(H) is an atom of $\mathscr{L}$ for every $H \in \mathscr{H}$, the whole model collapses into a classical probability model and the two conditionings (after appropriate normalization) degenerate into the classical Bayes' rule of conditioning. Identically if whenever $H \vdash L \vee L^*$ then either $H \vdash L$ or $H \vdash L^*$, then $\text{bel}_{\mathscr{L}}$ is a probability function (Smets 1993b).

4. Note : This derivation based on the probability of provability covers the cases generally considered by Dempster-Shafer theory (Dempster 1967), but not all those considered by the transferable belief model (TBM) where the probability measure on a hypothesis space is not necessarily assumed.

Dempster-Shafer theory has been criticized by the Bayesians as inappropriate: they claim that the conditioning by Dempster's rule of conditioning is inadequate. A strict Bayesian will claim the existence of a probability measure $P_{\mathscr{H} \times \mathscr{L}}$ on the product space $\mathscr{H} \times \mathscr{L}$ and ask for the application of Bayes' rule of conditioning on $P_{\mathscr{H} \times \mathscr{L}}$, and then the marginalization of the result on $\mathscr{L}$. Of course, the available information consists only on the marginalization of $P_{\mathscr{H} \times \mathscr{L}}$ on $\mathscr{H}$. The conditioning process cannot be achieved in general by lack of appropriate information. Only upper and lower conditional probabilities can be computed (Fagin and Halpern, 1990, Jaffray, 1992). Dempster's rule of conditioning is then inappropriate (Levi, 1983).

The only way to avoid the Bayesian criticisms consists in rejecting the probability measure on the product space $\mathscr{H} \times \mathscr{L}$ i.e. not accepting the Bayesian dogma that there exists a probability measure on ANY and EVERY space. Rejecting that probability measure on $\mathscr{H} \times \mathscr{L}$ is what is achieved explicitly in the Hints' model of Kohlas (and sometimes implicitly in Dempster-Shafer theory). In the TBM we even go further by not requiring the existence of any hypothesis space $\mathscr{H}$ and considering ONLY the $\mathscr{L}$ space by itself, cutting therefore all links with Dempster-Shafer model.

In the TBM, one considers only the basic belief assignment $m_\mathscr{G}$ and its related belief function $bel_\mathscr{G}$ and plausibility function $pl_\mathscr{G}$. No concept of some $\mathscr{H}$ space endowed with a probability measure is needed. The meaning of $m_\mathscr{G}(L)$ for $L \in \mathscr{G}$ is such that $m_\mathscr{G}(L)$ is the part of belief allocated to L and that could be allocated to any propositions L' that prove L if further information justifies such transfer. The Dempster's rule of conditioning is directly introduced as it is part of the overall description of the transferable belief model. The geometrical rule is derived if one ask for the proportion of the beliefs that support $L \in \mathscr{G}$ given they support $L^* \in \mathscr{G}$. Both rules have also been derived axiomatically in Smets (1992b) while looking for quantitative representations of credibility in general.

## 3. Jeffrey's rule applied to belief functions.

Let $(\Omega, \mathscr{A}, m_1)$ be a credibility space, where $\Omega$ is a finite set of worlds (the frame of discernment), $\mathscr{A}$ is a Boolean algebra of subsets built on $\Omega$ and $m_1$ is a basic belief assignment defined on $\mathscr{A}$. $m_1: \mathscr{A} \to [0,1]$ and its related evidential functions (the belief functions, possibility functions, communality functions... built from $m_1$) represent the agent initial belief about which world corresponds to the actual world.

Suppose the agent receives a new piece of evidence that tells him that his belief on the elements of a subalgebra $\mathscr{F}$ (with $\mathbb{B}$ its set of atoms) should be represented by the basic belief assignment $m_2: \mathscr{F} \to [0,1]$. $m_2$ is not defined on $\mathscr{A}$ but on $\mathscr{F}$. The agent wants to update his initial basic belief assignment $m_1$ into a new basic belief assignment $m_3$ on $\mathscr{A}$ that combines the information represented by $m_1$ and $m_2$. The first constraint is equivalent to constraint R1.

C1: $\forall X \in \mathscr{F}$, $bel_3(X) = bel_2(X)$

For every $A \in \mathscr{A}$, let $B(A) \in \mathscr{F}$ be the smallest element of $\mathscr{F}$ such that $A \subseteq B(A)$ and there is no other $B' \in \mathscr{F}$ such that $A \subseteq B' \subseteq B(A)$. $B(A)$ is the upper approximation of A in $\mathscr{F}$ in the sense of the rough sets theory (Pawlak 1982). Let $\mathscr{A}(A)$ be the set of $A \in \mathscr{A}$ that share the same $B(A)$.

The constraint C1 implies that the basic belief assignment $m_3$ is such that:

$$m_3(A) = \frac{c(A,B(A))}{\sum_{X:X \in \mathscr{A}(A)} c(X,B(A))} m_2(B(A)) \quad \forall A \in \mathscr{A},$$

where $c(A,B(A)) \geq 0$. The proof is based on the fact that the constraint C1 imposes that the basic belief masses $m_2(B(A))$ be allocated only to the elements $C \in \mathscr{A}(A)$. The way it is distributed among these C is arbitrary except the coefficients $c(A,B(A))$ must be positive so that $m_3$ is non negative.

The equivalent of constraint R2 is not immediate. In R2, we only had to consider the atoms of $\mathscr{F}$, as we were dealing with probability functions. Now, we must generalize the R2 requirement to every element of $\mathscr{F}$.

**a. Source-conditioning.**

C2F: $\forall B \in \mathbb{B}$, $\forall X,Y \in \mathscr{A}$ such that $X,Y \subseteq B$,

$$\frac{bel_3(X)}{bel_3(Y)} = \frac{bel_1(X \| B)}{bel_1(Y \| B)} \quad \text{if } bel_1(Y) > 0$$

and $bel_3(Y) = 0$ if $bel_1(Y) = 0$,

where $\|B$ in $bel_1(. \| B)$ denotes conditioning according to the geometrical rule of conditioning, in which case:

$$\frac{bel_1(X \| B)}{bel_1(Y \| B)} = \frac{bel_1(X)}{bel_1(Y)}.$$

The requirement C2F deals only with the elements of $\mathscr{A}$ that are subsets of the atoms of $\mathscr{F}$. For the other elements of $\mathscr{A}$, we propose:

C3F: $\forall X,Y \in \mathscr{A}$ such that $B(X) = B(Y)$,

$$\frac{\sum_{Z \subseteq X, B(Z)=B(X)} m_3(Z)}{\sum_{Z \subseteq Y, B(Z)=B(Y)} m_3(Z)} = \frac{\sum_{Z \subseteq X, B(Z)=B(X)} m_1(Z \| B(X))}{\sum_{Z \subseteq Y, B(\ =B(Y)} m_1(Z \| B(Y))}$$

(where the left denominator is zero is the right one is zero).

C2F is the particular case of C3F that matches the R2 requirement. C3F has to be added as belief functions must be defined on all the elements of $\mathscr{A}$, not only on the atoms of $\mathscr{A}$.

It is straightforward to show that C1F and C3F are satisfied iff:

$$m_3(A) = \frac{m_1(A)}{\sum_{X:X \in \mathscr{A}(A)} m_1(X)} m_2(B(A)) \quad \forall A \in \mathscr{A} \text{ if the}$$

denominator is positive and $m_3(A) = 0$ otherwise.

We propose to call this rule the **Jeffrey geometric rule of conditioning**. Indeed, if there is only one $B \in \mathscr{F}$, such that $m_2(B) = 1$, then $m_3$ is the basic belief assignment obtained from $m_1$ by conditioning on B with the geometric rule of conditioning.

**b. Data-conditioning.**

The constraints corresponding to C2F and C3F are:

**C2R:** $\forall B \in \mathbb{B}, \forall X, Y \in \mathcal{A}$ such that $X, Y \subseteq B$,

$$\frac{bel_3(X)}{bel_3(Y)} = \frac{bel_1(X|B)}{bel_1(Y|B)} \quad \text{if } bel_1(Y|B) > 0$$

and $\quad bel_3(Y) = 0 \quad \text{if } bel_1(Y|B) = 0$,

where $|B$ in $bel_1(.|B)$ denotes conditioning according to Dempster's rule of conditioning.

**C3R:** $\forall X, Y \in \mathcal{A}$ such that $B(X) = B(Y)$,

$$\frac{\sum_{Z \subseteq X, B(Z)=B(X)} m_3(Z)}{\sum_{Z \subseteq Y, B(Z)=B(Y)} m_3(Z)} = \frac{\sum_{Z \subseteq X, B(Z)=B(X)} m_1(Z|B(X))}{\sum_{Z \subseteq Y, B(Z)=B(Y)} m_1(Z|B(Y))}$$

(where the left denominator is zero is the right one is zero).

It is straightforward to show that C1R and C3R are satisfied iff:

$$m_3(A) = \frac{m_1(A|B(A))}{\sum_{X: X \in \mathcal{A}(A)} m_1(X|B(A))} m_2(B(A)) \quad \forall A \in \mathcal{A}$$

if the denominator is positive and $m_3(A) = 0$ otherwise.

We propose to call this rule the **Jeffrey-Dempster rule of conditioning**. Indeed, if there is only one $B \in \mathcal{B}$, such that $m_2(B) = 1$, then $m_3$ is the basic belief assignment obtained from $m_1$ by conditioning on B with Dempster's rule of conditioning.

The justification of C2F and C2R are quite straightforward. They are the same as with Jeffrey original case. In every atom of $\mathcal{B}$, the ratio of the probabilities $P_3$ given to two elements of $B \in \mathbb{B}$ is equal to the ratio of the conditional probabilities $P_1$ given to these elements after conditioning on B. For the belief function generalization, we have the choice between conditioning by the geometrical rule of conditioning or the unnormalized rule of conditioning, therefore the two cases. Unfortunately C2R and C2F are not sufficient to derive both Jeffrey's rules. The C3F and C3R are proposed as the generalization of C2F and C2R to those elements of $\mathcal{A}$ that intersect *several* elements of $\mathcal{B}$.

The C3F and C3R rules are less obvious. Their meaning is as follows. Each sum is the part of belief related to the basic belief masses given to the elements of $\mathcal{A}$ that were not yet allocated to some of its subsets by requirement C1. That their ratio should be equal is the extension of the requirement used in C2F and C2R. Indeed, C3F and C3R correspond to the case C2F and C2R when $B(X) \in \mathbb{B}$.

It could of course be arguable but they seem to be the most natural requirements to propose to generalize the C2F and C2R requirements.

## 4. Conclusions.

Shafer (1981) has studied Jeffrey's rule. He considers its generalization can be found in Dempster's rule of combination. Let $bel_{12} = bel_1 \oplus bel_2$. Shafer notes that:

$$\forall B \in \mathbb{B}, \forall A \subseteq B, \quad bel_{12}(A|B) = bel_1(A|B)$$

He considers that this relation fits with Jeffrey's aims. His proposal does not fit with C1 and we feel C1 is more in the spirit of Jeffrey's updating then Shafer's proposal. As noted by Dubois and Prade (1986, 1992), the characteristic of Jeffrey's rule is its asymmetry. $bel_2$ is 'more important' than $bel_1$ in that $bel_3$ must be equal to $bel_2$ for all $X \in \mathcal{B}$ whatever $bel_1$ (requirement C1). In the limiting case where $\mathcal{B} = \mathcal{A}$, $bel_3 = bel_2$ and $bel_1$ is completely ignored. Of course Dempster's rule of combination does not satisfy this idea. Fundamentally, it is a symmetrical rule.

In Dubois and Prade (1991b, 1992), the authors suggest another generalization.

$$bel_3(A) = \sum_{B: B \in \mathcal{B}} \frac{bel_1(A|B)}{pl_1(B)} m_2(B) \quad \forall A \in \mathcal{A} \quad (4.1)$$

The normalization factor $pl_1(B)$ is essential in their formulation. If it were not introduced, the rule would be equivalent to the unnormalized Dempster's rule of combination what is exactly what they want to avoid. Unfortunately, their proposal fails also to satisfy C1.

Ichihashi and Tanaka (1989) have suggested the following three generalizations of Jeffrey's rule (se also Wagner 1992):

$$bel_3(A) = \sum_{B: B \in \mathcal{B}} f(A, B) m_2(B) \quad \forall A \in \mathcal{A}$$

where $\quad f(A, B) = \dfrac{bel_1(A \vee \overline{B}) - bel_1(\overline{B})}{pl_1(B)}$

$f(A, B) = \dfrac{bel_1(A \wedge B)}{bel_1(B)}$

$f(A, B) = \dfrac{bel_1(A) - bel_1(A \wedge \overline{B})}{pl_1(B)}$

These proposals fail to satisfy C1 (the first is 4.1).

Wagner (1992) solves the case where $bel_1$ is a probability function in a context of upper and lower probabilities. His solution is covered by our solutions.

In conclusion, we have presented what we feel is the fundamental meaning of Jeffrey's rule and how it generalizes within the belief functions framework. The applicability of the generalized Jeffrey's rules of conditioning resides in the case where the updating information corresponds to some constraints that must be satisfied by the updated belief function after combination of the updating information with the initial state of belief, but where the updating information induces a belief function only on a subalgebra of the algebra on which the initial belief function is defined. It does not covered the case of an updating on an unreliable information (what Dubois and Prade claim as being appropriately modeled by 4.1). It requires that the updating information characterized by $bel_2$ be such that our revised belief $bel_3$ must be equal to $bel_2$ wherever $bel_2$ is defined. Should $bel_2$ be defined on the same algebra as $bel_1$ (i.e., $\mathscr{A} = \mathscr{B}$), the updating would result into adopting $bel_2$ as the revised belief. That requirement fits with the idea of a revision by readaptation (or correction) of $bel_1$ by $bel_2$. In a certain sense, it satisfies the 'success rule' described for revision ($A \in K^*_A$, Gardenfors 1988).

The nature of the atoms of $\mathscr{B}$ is important in order to apply Jeffrey's rules. In practice, $bel_2$ is provided by a source of evidence that specifies $bel_2$ values on some elements of $\mathscr{A}$. $\mathscr{B}$ is then the coarsest Boolean subalgebra of $\mathscr{A}$ that contains all the elements of $\mathscr{A}$ on which $bel_2$ is known.

## References.